\documentclass[sigconf,natbib=true,anonymous=false]{acmart}

\usepackage{graphicx}
\usepackage{array}
\usepackage{makecell}
\usepackage{multirow}
\usepackage{booktabs}
\usepackage{amsmath}
\usepackage{dblfloatfix}
\usepackage{soul}
\usepackage{subfig}

\AtBeginDocument{%
  \providecommand\BibTeX{{%
    \normalfont B\kern-0.5em{\scshape i\kern-0.25em b}\kern-0.8em\TeX}}}

\setcopyright{iw3c2w3}
\copyrightyear{2023}
\acmYear{2023}
\acmDOI{XXXXXXX.XXXXXXX}

\acmConference[Preprint]{Arxiv}{2023}{Arxiv}
\acmPrice{15.00}
\acmISBN{978-1-4503-XXXX-X/18/06}

\begin{document}

\title{Class Anchor Margin Loss for Content-Based Image Retrieval}

\author{Alexandru Ghi\c{t}\u{a}}
\affiliation{%
  \institution{Department of Computer Science, University of Bucharest}
  \streetaddress{14 Academiei}
  \city{Bucharest}
  \country{Romania}
  \postcode{010014}
}
\email{alexandru.ghita94@gmail.com}

\author{Radu Tudor Ionescu}
\authornote{Corresponding author.}
\affiliation{%
  \institution{Department of Computer Science, University of Bucharest}
  \streetaddress{14 Academiei}
  \city{Bucharest}
  \country{Romania}
  \postcode{010014}
}
\email{raducu.ionescu@gmail.com}

\renewcommand{\shortauthors}{Ghi\c{t}\u{a} et al.}

\begin{abstract}
The performance of neural networks in content-based image retrieval (CBIR) is highly influenced by the chosen loss (objective) function. The majority of objective functions for neural models can be divided into metric learning and statistical learning. Metric learning approaches require a pair mining strategy that often lacks efficiency, while statistical learning approaches are not generating highly compact features due to their indirect feature optimization. To this end, we propose a novel repeller-attractor loss that falls in the metric learning paradigm, yet directly optimizes for the $L_{2}$ metric without the need of generating pairs. Our loss is formed of three components. One leading objective ensures that the learned features are attracted to each designated learnable class anchor. The second loss component regulates the anchors and forces them to be separable by a margin, while the third objective ensures that the anchors do not collapse to zero. Furthermore, we develop a more efficient two-stage retrieval system by harnessing the learned class anchors during the first stage of the retrieval process, eliminating the need of comparing the query with every image in the database. We establish a set of four datasets (CIFAR-100, Food-101, SVHN, and Tiny ImageNet) and evaluate the proposed objective in the context of few-shot and full-set training on the CBIR task, by using both convolutional and transformer architectures. Compared to existing objective functions, our empirical evidence shows that the proposed objective is generating superior and more consistent results. 
\end{abstract}

\ccsdesc[500]{Computing methodologies~Machine learning}
\ccsdesc[500]{Computing methodologies~Ranking}
\ccsdesc[500]{Computing methodologies~Supervised learning by classification}

\keywords{contrastive learning, contrastive loss, class anchors, content-based image retrieval, object retrieval}


\maketitle

\section{Introduction}

Content-based image retrieval (CBIR) is a challenging task that aims at retrieving images from a database that are most similar to a query image. The state-of-the-art image retrieval systems addressing the aforementioned task are currently based on deep neural networks \cite{Cao-ECCV-2020,Dubey-TCSVT-2021,Lee-CVPR-2022,Radenovic-TPAMI-2019,Revaud-ICCV-2019,Wu-ICCV-2021}. Although neural models obtained impressive performance levels compared to handcrafted CBIR models \cite{Philbin-CVPR-2007}, one of the main challenges that remain to be solved in training neural networks for retrieval problems is the choice of the objective function. Indeed, a well-chosen objective function should enhance the discriminative power of the learned embeddings, i.e.~the resulting features should exhibit small differences for images representing the same object and large differences for images representing different objects. This would implicitly make the embeddings more suitable for image retrieval. 

The majority of loss functions used to optimize neural models can be separated into two major categories: statistical learning and metric learning. The most popular category is represented by statistical learning, which includes objective functions such as the cross-entropy loss \cite{Murphy-MIT-2012} or the hinge loss \cite{Cortes-ML-1995}. The optimization problem is usually based on minimizing a particular probability distribution. Consequently, objective functions in this category achieve the desired properties as an indirect effect during optimization. Hence, objective functions based on statistical learning are more suitable for modeling multi-class classification tasks rather than retrieval tasks. The second category corresponds to metric learning, which comprises objective functions such as contrastive loss \cite{Hadsell-CVPR-2006}, triplet loss \cite{Schroff-CVPR-2015}, and quadruplet loss \cite{Chen-CVPR-2017}. Objective functions in this category operate directly in the embedding space and optimize for the desired distance metric. However, they usually require forming tuples of examples to compute the loss, which can be a costly step in terms of training time. To reduce the extra time required to build tuples, researchers resorted to hard example mining schemes \cite{Georgescu-ICPR-2021,Georgescu-MVA-2022,Harwood-ICCV-2017,Schroff-CVPR-2015,Suh-CVPR-2019, Wu-ICCV-2017}. Still, statistical learning remains more time efficient.

In this context, we propose a novel repeller-attractor objective function that falls in the metric learning paradigm. Our loss directly optimizes for the $L_{2}$ metric without needing to generate pairs, thus alleviating the necessity of performing the costly hard example mining. The proposed objective function achieves this through three interacting components, which are expressed with respect to a set of learnable embeddings, each representing a class anchor. The leading loss function ensures that the data embeddings are attracted to the designated class anchor. The second loss function regulates the anchors and forces them to be separable by a margin, while the third objective ensures that the anchors do not collapse to the origin. In addition, we propose a two-stage retrieval method that compares the query embedding with the class anchors in the first stage, then continues by comparing the query embedding with image embeddings assigned to the nearest class anchors. By harnessing the learned class anchors, our retrieval process becomes more efficient and effective than the brute-force search.

We carry out experiments on four datasets (CIFAR-100 \cite{Krizhevsky-TECHREP-2009}, Food-101 \cite{Lukas-ECCV-2014}, SVHN \cite{Netzer-DLUFL-2011}, and Tiny ImageNet \cite{Russakovsky-IJCV-2015}) to compare the proposed loss function against representative statistical and deep metric learning objectives. We evaluate the objectives in the context of few-shot and full-set training on the CBIR task, by using both convolutional and transformer architectures, such as residual networks (ResNets) \cite{He-CVPR-2016} and shifted windows (Swin) transformers \cite{Liu-ICCV-2021}. Moreover, we test the proposed losses on various embedding space dimensions, ranging from $32$ to $2048$. Compared to existing loss functions, our empirical results show that the proposed objective is generating higher and more consistent performance levels across the considered evaluation scenarios. Furthermore, we conduct an ablation study to demonstrate the influence of each loss component on the overall performance.

In summary, our contribution is threefold:
\begin{itemize}
\item We introduce a novel repeller-attractor objective function that directly optimizes for the $L_{2}$ metric, alleviating the need to generate pairs via hard example mining or alternative mining strategies.
\item We propose a two-stage retrieval system that leverages the use of the learned class anchors in the first stage of the retrieval process, leading to significant speed and performance gains.
\item We conduct comprehensive experiments to compare the proposed loss function with popular loss choices in multiple evaluation scenarios.
\end{itemize}

\section{Related Work}

As related work, we refer to studies introducing new loss functions, which are related to our first contribution, and new content-based image retrieval methods, which are connected to our second contribution.

\noindent
\textbf{Loss functions.}
The problem of generating features that are tightly clustered together for images representing the same objects and far apart for images representing distinct objects is a challenging task usually pursued in the context of CBIR. For retrieval systems based on neural networks, the choice of the objective function is the most important factor determining the geometry of the resulting feature space \cite{Tang-CIKM-2022}. We hereby discuss related work proposing various loss functions aimed at generating effective embedding spaces.

Metric learning objective functions directly optimize a desired metric and are usually based on pairs or tuples of known data samples~\cite{Sohn-NIPS-2016, r:local_features_desc_cnn,r:tuplet_margin_loss,Chen-CVPR-2017}. 
One of the earliest works on metric learning proposed the contrastive loss~\cite{Hadsell-CVPR-2006}, which was introduced as a method to reduce the dimensionality of the input space, while preserving the separation of feature clusters. The idea behind contrastive loss is to generate an attractor-repeller system that is trained on positive and negative pairs generated from the available data. The repelling can happen only if the distance between a negative pair is smaller than a margin $m$. In the context of face identification, another successful metric learning approach is triplet loss~\cite{Schroff-CVPR-2015}. It obtains the desired properties of the embedding space by generating triplets of anchor, positive and negative examples. Each image in a batch is considered an anchor, while positive and negative examples are selected online from the batch. For each triplet, the proposed objective enforces the distance between the anchor and the positive example to be larger than the distance between the anchor and the negative example, by a margin $m$. Other approaches introduced objectives that directly optimize the AUC \cite{Gajic-CVPR-2022}, recall \cite{Patel-CVPR-2022} or AP \cite{Revaud-ICCV-2019}. The main issues when optimizing with loss functions based on metric learning are the usually slow convergence \cite{Sohn-NIPS-2016} and the difficulty of generating useful example pairs or tuples \cite{Suh-CVPR-2019}. In contrast, our method does not require mining strategies and, as shown in the experiments, it converges much faster than competing losses.

The usual example mining strategies are hard, semi-hard, and online negative mining \cite{Schroff-CVPR-2015, Wu-ICCV-2017}. In hard negative mining, for each anchor image, we need to construct pairs with the farthest positive example and the closest negative example. This adds an extra computational step at the beginning of every training epoch, extending the training time. Similar problems arise in the context of semi-hard negative mining, while the difference consists in the mode in which the negatives are sampled. Instead of generating pairs with the farthest example, one can choose a negative example that is slightly closer than a positive sample, thus balancing hardness and stability. A more efficient approach is online negative mining, where negative examples are sampled during training in each batch. In this approach, the pair difficulty adjusts while the model is training, thus leading to a more efficient strategy, but the main disadvantage is that the resulting pairs may not be the most challenging for the model, due to the randomness of samples in a batch.

Statistical learning objective functions indirectly optimize the learned features of the neural network. Popular objectives are based on some variation of the cross-entropy loss~\cite{r:arcface, r:sphere_face, r:cosface, r:additive_margin_softmax, r:large_margin_softmax_loss, Elezi-TPAMI-2022} or the cosine loss~\cite{r:cosine_loss}. By optimizing such functions, the model is forced to generate features that are close to the direction of the class center. For example, ArcFace~\cite{r:arcface} reduces the optimization space to an $n$-dimensional hypersphere by normalizing both the embedding generated by the encoder, and the corresponding class weight from the classification head, using their Euclidean norm. An additive penalty term is introduced, and with the proposed normalization, the optimization is performed on the angle between each feature vector and the corresponding class center.

Hybrid objective functions promise to obtain better embeddings by minimizing a statistical objective function in conjunction with a metric learning objective \cite{Min-TMM-2020,r:sup_con_loss}. For example, Center Loss \cite{r:center_loss} minimizes the intra-class distances of the learned features by using cross-entropy in conjunction with an attractor for each sample to its corresponding class center. During training, the class centers are updated to become the mean of the feature vectors for every class seen in a batch. Another approach \cite{r:cac_loss} similar to Center Loss~\cite{r:center_loss} is based on predefined evenly distributed class centers. 
A more complex approach \cite{Cao-ECCV-2020} is to combine the standard cross-entropy with a cosine classifier and a mean squared error regression term to jointly enhance  global and local features.


Both contrastive and triplet loss objectives suffer from the need of employing pair mining strategies, but in our case, mining strategies are not required. The positive pairs are built online for each batch, between each input feature vector and the dedicated class anchor, the number of positive pairs thus being equal to the number of examples. The negative pairs are constructed only between the class centers, thus alleviating the need of searching for a good negative mining strategy, while also significantly reducing the number of negative pairs. To the best of our knowledge, there are no alternative objective functions for CBIR that use dedicated self-repelling learnable class anchors acting as attraction poles for feature vectors belonging to the respective classes.

\noindent
\textbf{Content-based image retrieval methods.}
CBIR systems are aimed at finding similar images with a given query image, matching the images based on the similarity of their scenes or the contained objects. Images are encoded using a descriptor (or encoder), and a system is used to sort a database of encoded images based on a similarity measure between queries and images.
In the context of content-based image retrieval, there are two types of image descriptors. On the one hand, there are general descriptors \cite{Radenovic-TPAMI-2019}, where a whole image is represented as a feature vector. On the other hand, there are local descriptors \cite{Wu-ICCV-2021} where portions of an image are represented as feature vectors. Hybrid descriptors \cite{Cao-ECCV-2020} are also used to combine both global and local features. To improve the quality of the  results retrieved by learned global descriptors, an additional verification step is often employed. This step is meant to re-rank the retrieved images by a precise evaluation of each candidate image \cite{Polley-CIKM-2022}. The re-ranking step is usually performed with the help of an additional system, and in some cases, it can be directly integrated with the general descriptor \cite{Lee-CVPR-2022}.
In the CBIR task, one can search for visually similar images as a whole, or search for images that contain similar regions \cite{Philbin-CVPR-2007} of a query image. In this work, we focus on building global descriptors that match whole images. Further approaches based on metric learning, statistical learning, or hand-engineered features are discussed in the recent survey of Dubey et al.~\cite{Dubey-TCSVT-2021}. Different from other CBIR methods, we propose a novel two-stage retrieval system that leverages the use of the class anchors learned through the proposed loss function to make the retrieval process more efficient and effective.

\section{Method} \label{s:method}

\noindent
\textbf{Overview.}
Our objective function consists of three components, each playing a different role in obtaining a discriminative embedding space. All three loss components are formulated with respect to a set of learnable class anchors (centroids). The first loss component acts as an attraction force between each input embedding and its corresponding class anchor. Its main role is to draw the embeddings representing the same object to the corresponding centroid, thus creating embedding clusters of similar images. Each center can be seen as a magnet with a positive charge and its associated embeddings as magnets with negative charges, thus creating attraction forces between anchors and data samples of the same class. The second loss component acts as a repelling force between class anchors. In this case, the class centers can be seen as magnets with similar charges. If brought together, they will repel each other, and if they lie at a certain distance, the repelling force stops. The last component acts similarly to the second one, with the difference that an additional magnet is introduced and fixed at the origin of the embedding space. Its main effect is to push the class centers away from the origin.

\noindent
\textbf{Notations.}
Let $\mathbf{x}_i \in \mathbb{R}^{h\times w\times c}$ be an input image and $y_i \in \mathbb{N}$ its associated class label, $\forall i \in \{1, 2,...,l\}$. We aim to optimize a neural encoder $f_{\theta}$ which is parameterized by the learnable weights $\theta$ to produce a discriminative embedding space. Let $\mathbf{e}_i \in \mathbb{R}^{n}$ be the $n$-dimensional embedding vector of the input image $\mathbf{x}_i$ generated by $f_{\theta}$, i.e.~$\mathbf{e}_i = f_{\theta}(\mathbf{x}_i)$. In order to employ our novel loss function, we need to introduce a set of learnable class anchors $C=\lbrace \mathbf{c}_1, \mathbf{c}_2 , ..., \mathbf{c}_t \rbrace$, where $\mathbf{c}_j \in \mathbb{R}^n$ resides in the embedding space of $f_{\theta}$, and $t$ is the total number of classes.

\begin{figure}[t]
\centerline{\includegraphics[width=1.0\linewidth]{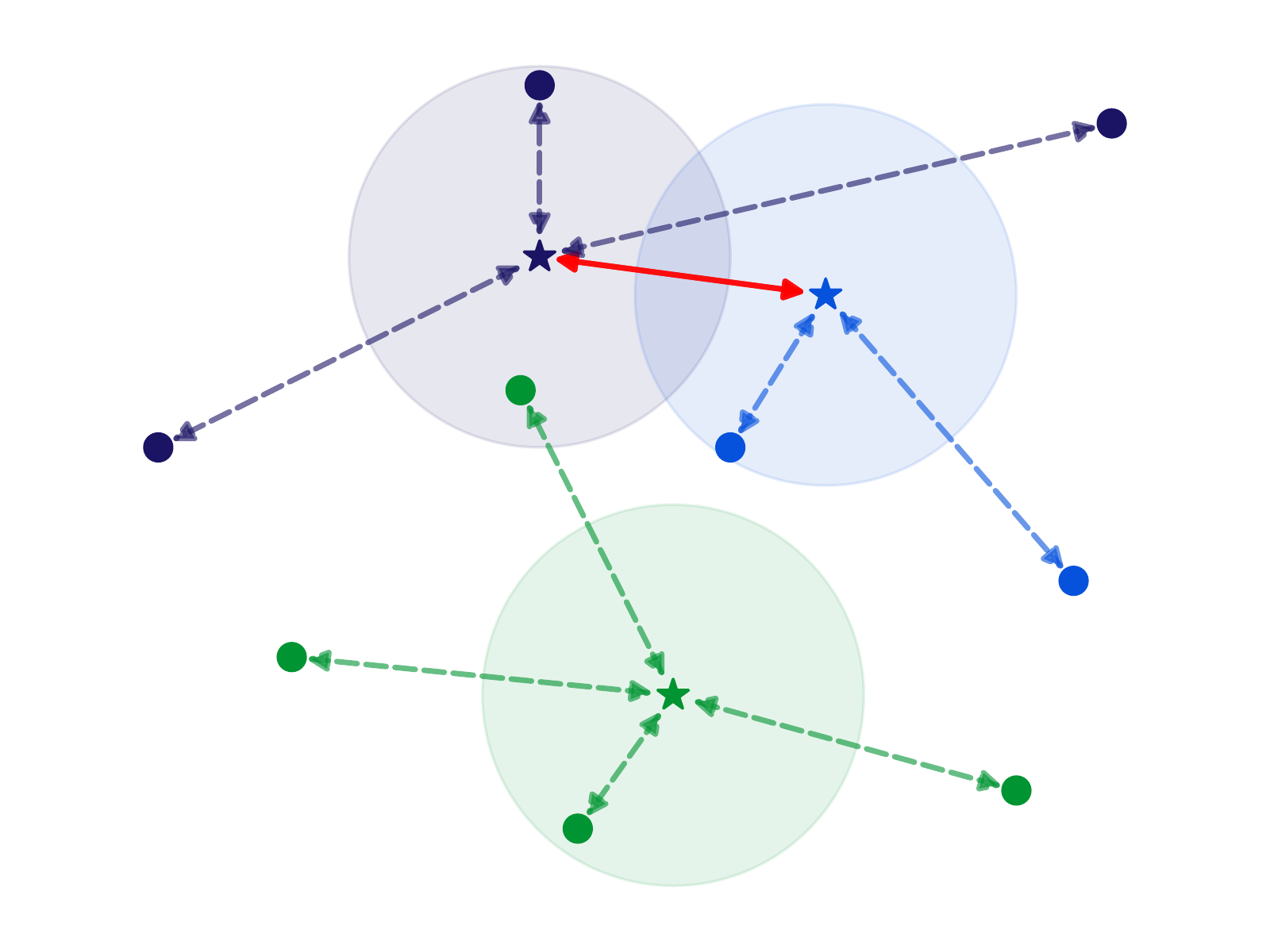}}
\caption{An example showing the behavior of the attractor-repeller loss components for three classes. The stars represent the class anchors $C$. Faded circles around class anchors represent the sphere of radius $m$ around each anchor. Solid circles represent feature vectors generated by the encoder $f_{\theta}$.
Dashed arrows between feature vectors and class anchors represent the attraction of the force generated by the attractor $\mathcal{L}_{A}$. The solid red arrow between class anchors represents the repelling force generated by the repeller $\mathcal{L}_{R}$. Best viewed in color.}
\label{f:attractor_repeller}
\end{figure}

\noindent
\textbf{Loss components.}
With the above notations, we can now formally define our first component, the attractor loss $\mathcal{L}_A$, as follows:
\begin{equation}
    \mathcal{L}_{A}(\mathbf{x}_i, C) = \frac{1}{2} \| \mathbf{e}_{i} - \mathbf{c}_{y_{i}} \|_{2}^{2}.
\label{attractor}
\end{equation}
The main goal of this component of the objective is to cluster feature vectors as close as possible to their designated class anchor by minimizing the distance between $\mathbf{e}_{i}$ and the corresponding class anchor $\mathbf{c}_{y_{i}}$. Its effect is to enforce low intra-class variance. However, obtaining low intra-class variance is only a part of what we aim to achieve. The first objective has little influence over the inter-class similarity, reducing it only indirectly. Therefore, another objective is required to repel samples from different classes. As such, we introduce the repeller loss $\mathcal{L}_R$, which is defined as follows:
\begin{equation}
    \mathcal{L}_R(C) = \frac{1}{2} \sum_{y, y' \in Y, y \neq y'} \left\{ \max \left( 0, 2\cdot m -  \| \mathbf{c}_{y} - \mathbf{c}_{y'} \| \right) \right\}^{2},
\label{repeller}
\end{equation}
where $y$ and $y'$ are two distinct labels from the set of ground-truth labels $Y$, and $m > 0$ is a margin representing the radius of an $n$-dimensional sphere around each anchor, in which no other anchor should lie. The goal of this component is to push anchors away from each other during training to ensure high inter-class distances. The margin $m$ is used to limit the repelling force to an acceptable margin value. If we do not set a maximum margin, then the repelling force can push the anchors too far apart, and the encoder could struggle to learn features that satisfy the attractor loss defined in Eq.~\eqref{attractor}. 

\begin{figure}[t]
\centerline{\includegraphics[width=0.8\linewidth]{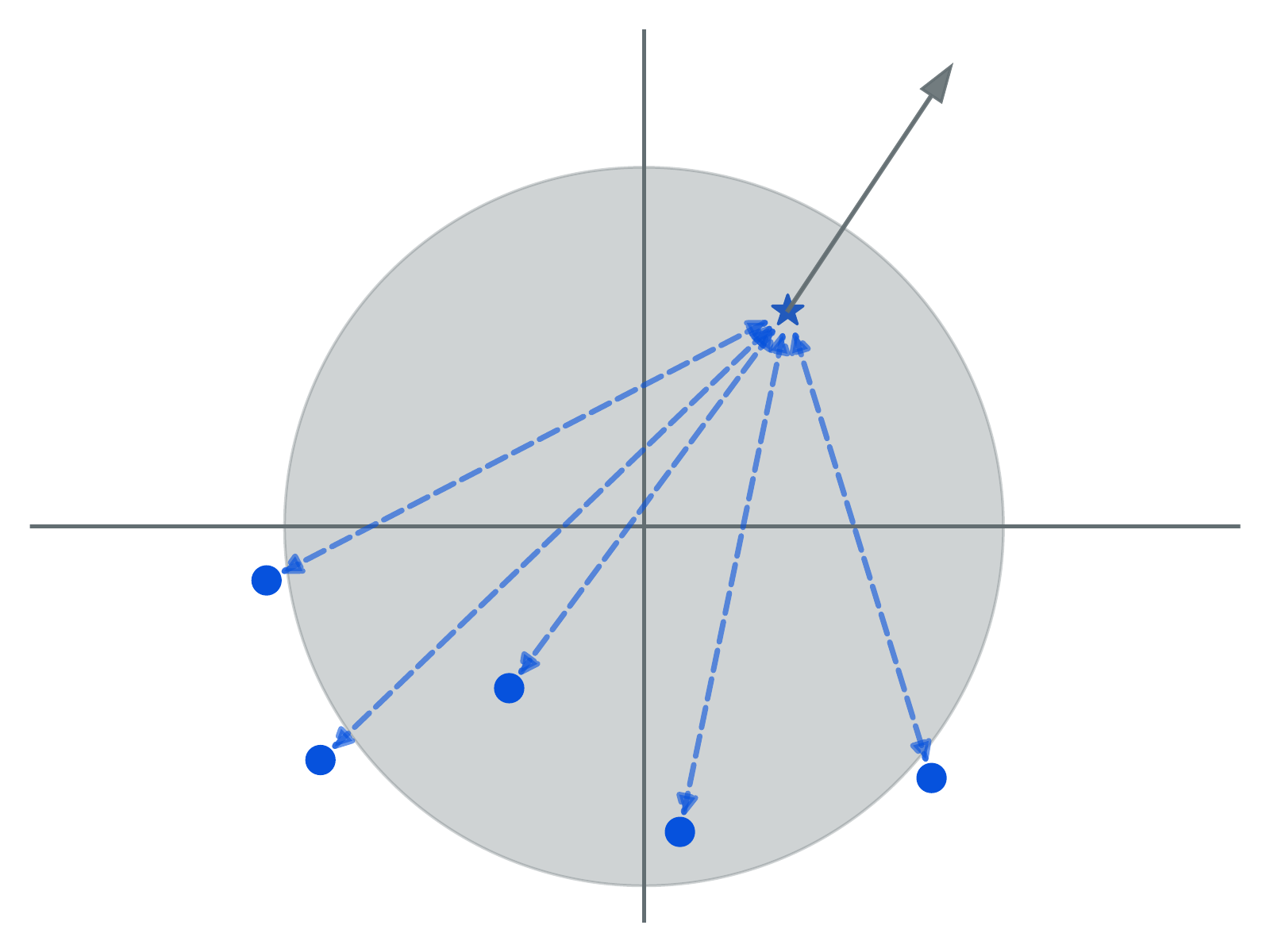}}
\caption{Contribution of the minimum norm loss $\mathcal{L}_{N}$ imposed on the class anchors. The blue star represents a class anchor. Solid circles represent embedding vectors generated by the encoder $f_{\theta}$. Dashed arrows represent the attraction force generated by the attractor $\mathcal{L}_{A}$. The solid gray line represents the direction in which the anchor is pushed away from the origin due to the component $\mathcal{L}_{N}$. Best viewed in color.}
\label{f:min_anchors_norm}
\end{figure}

\begin{figure*}[t]
\centerline{
\includegraphics[height=2.3in]{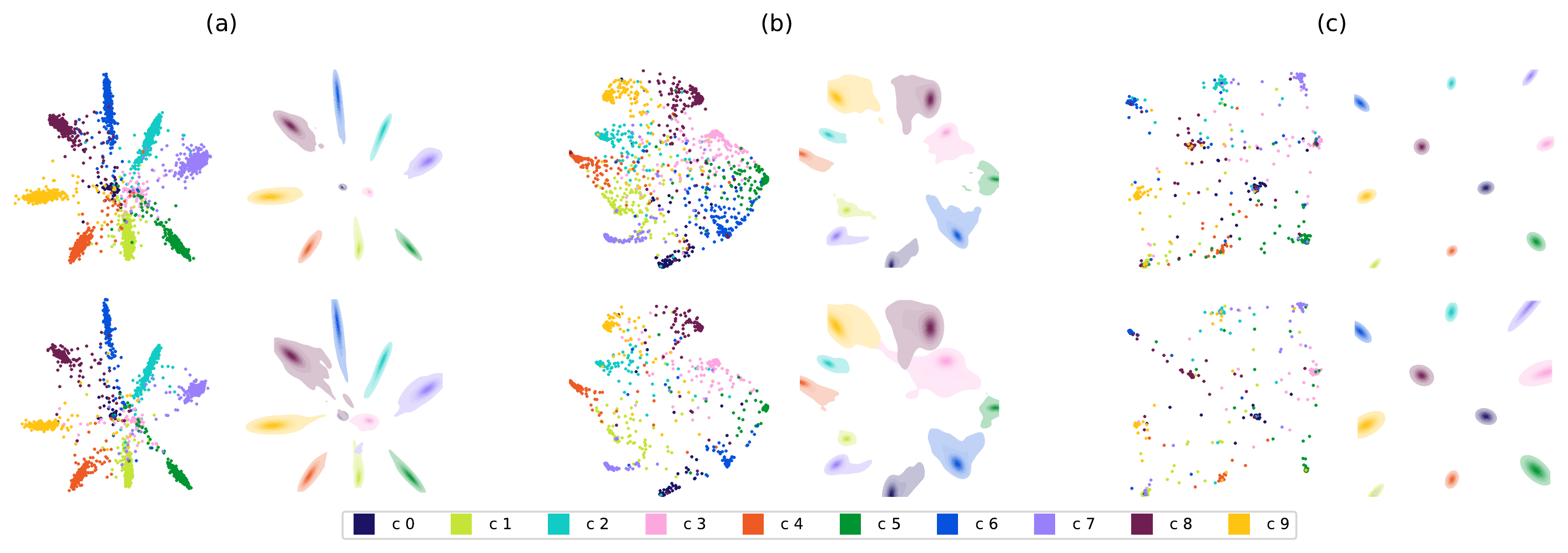}}
\caption{Embedding vectors (left) and their distribution (right) for a ResNet-18 model modified to output 2D features and trained with $(a)$ cross-entropy loss, $(b)$ contrastive loss and $(c)$ class anchor margin loss (ours). The top row represents training embeddings from the SVHN dataset, while the bottom row shows test embeddings. Best viewed in color.}
\label{f:embeds_dist_plot}
\end{figure*}

A toy example of the attractor-repeller mechanism is depicted in Figure~\ref{f:attractor_repeller}. Notice how the optimization based on the attractor-repeller objective tends to pull data samples from the same class together (due to the attractor), and push samples from different classes away (due to the repeller). However, when the training begins, all data samples start from a location close to the origin of the embedding space, essentially having a strong tendency to pull the class anchors to the origin. To ensure that the anchors do not collapse to the origin (as observed in some of our preliminary experiments), we introduce an additional objective that imposes a minimum norm on the class anchors. The minimum norm loss $\mathcal{L}_N$ is defined as:
\begin{equation}
    \mathcal{L}_{N}(C) = \frac{1}{2} \sum_{y \in Y} \left\{ \max \left( 0, p - \| \mathbf{c}_{y} \| \right) \right\}^{2},
\label{anchorsminnorm}
\end{equation}
where $p$ is the minimum norm that each anchor must have. This objective contributes to our full loss function as long as at least one class anchor is within a distance of $p$ from the origin. Figure~\ref{f:min_anchors_norm} provides a visual interpretation of the effect induced by the minimum norm loss. Notice how the depicted class anchor is pushed away from the origin (due to the minimum norm loss), while the data samples belonging to the respective class move along with their anchor (due to the attractor loss).

Assembling the three loss components presented above into a single objective leads to the proposed class anchor margin (CAM) loss $\mathcal{L}_{\scriptsize{\mbox{CAM}}}$, which is formally defined as follows:
\begin{equation}
\mathcal{L}_{\scriptsize{\mbox{CAM}}}(\mathbf{x}, C) = \mathcal{L}_{A}(\mathbf{x},C) + \mathcal{L}_{R}(C) + \mathcal{L}_{N}(C).
\label{finalloss}
\end{equation}
Notice that only $\mathcal{L}_A$ is directly influenced by the training examples, while $\mathcal{L}_R$ and $\mathcal{L}_N$ operate only on the class anchors. Hence, negative mining strategies are not required at all. 

\noindent
\textbf{Gradients of the proposed loss.}
We emphasize that the $L_{2}$ norm is differentiable. Moreover, the updates for the weights $\theta$ of the encoder $f_{\theta}$ are provided only by the attractor loss $\mathcal{L}_{A}$. Hence, the weight updates (gradients of $\mathcal{L}_{\scriptsize{\mbox{CAM}}}$ with respect to $\theta$) for some data sample $\mathbf{x}_i$ are computed as follows:
\begin{equation}
    \frac{\partial \mathcal{L}_{\scriptsize{\mbox{CAM}}}(\mathbf{x}_i,C)}{\partial \theta} = \frac{\partial \mathcal{L}_{A}(\mathbf{x}_i,C)}{\partial \theta} = \left( f_{\theta}(\mathbf{x}_i) - \mathbf{c}_{y_{i}}\right) \cdot\frac{\partial f_{\theta}(\mathbf{x}_i)^{T}}{\partial \theta}.
\end{equation}

The class anchors receive updates from all three loss components. For a class anchor $\mathbf{c}_{y}$, the update received from the component $\mathcal{L}_{A}$ of the joint objective is given by:
\begin{equation}
    \frac{\partial \mathcal{L}_{A}(\mathbf{x},C)}{\partial \mathbf{c}_{y}} = \mathbf{c}_{y} - f_{\theta}(\mathbf{x}).
    \label{eq:partial_attractor_centers}
\end{equation}

The contribution of the repeller to a class anchor $\mathbf{c}_{y}$ is null when the $m$-margin sphere of the respective anchor is not intersecting another class anchor sphere, i.e.:
\begin{equation}
\frac{\partial \mathcal{L}_{R}(C)}{\partial C} = 0,
\end{equation}
when $2\cdot m -\| \mathbf{c}_{y} - \mathbf{c}_{y'} \| > 0$, $\forall y, y' \in Y, y \neq y'$. 

To simply the notation, let $d = \| \mathbf{c}_{y} - \mathbf{c}_{y'} \|$ be the distance between $\mathbf{c}_{y}$ and another class center $\mathbf{c}_{y'}$. The update for a class center $\mathbf{c}_{y}$ of the repeller is given by:
\begin{equation}
    \frac{\partial \mathcal{L}_{R}(C)}{\partial \mathbf{c}_{y}} =\! \sum_{y, y' \in Y, y \neq y'}\!\delta(d<2\!\cdot\!m) \cdot \frac{-(2\!\cdot\!m - d)\!\cdot\!(\mathbf{c}_{y} - \mathbf{c}_{y'})}{d},
    \label{eq:partial_repeller_centers}
\end{equation}
where $\delta(*) = 1$ when $*$ is satisfied, and $0$ otherwise.
Similarly, the contribution of the minimum norm loss is given by:
\begin{equation}
    \frac{\partial \mathcal{L}_{N}(C)}{\partial \mathbf{c}_{y}} = \sum_{y \in Y} \delta\left(\| \mathbf{c}_{y} \|<p\right) \cdot \left(1 -  \frac{p}{\| \mathbf{c}_{y} \|} \right) \cdot \mathbf{c}_{y}.
    \label{eq:partial_min_anchors_norm_centers}
\end{equation}

The final update of the learnable class centers $C$ is given by summing the gradients given in Eq.~\eqref{eq:partial_attractor_centers}, Eq.~\eqref{eq:partial_repeller_centers} and Eq.~\eqref{eq:partial_min_anchors_norm_centers}:
\begin{equation}
    \frac{\partial \mathcal{L}(\mathbf{x}, C)}{\partial \mathbf{c}_{y}} = \frac{\partial \mathcal{L}_{A}(\mathbf{x},C)}{\partial \mathbf{c}_{y}} + \frac{\partial \mathcal{L}_{R}(C)}{\partial \mathbf{c}_{y}} + \frac{\partial \mathcal{L}_{N}(C)}{\partial \mathbf{c}_{y}}. 
\end{equation}

\noindent
\textbf{Fast retrieval via two-stage system.}
During inference, we can employ the $L_{2}$ measure between query and database embeddings to retrieve images similar to the query. Aside from the brute-force search that compares the query with every embedding in the database, we can harness the geometric properties of the resulting embedding space and use the class anchors $C$ to improve the speed of the retrieval system. Instead of computing the $L_{2}$ distances between a query feature vector and all the embedding vectors stored in the database, we propose to employ a two-stage system. In the first stage, distances are computed between the query feature vector and each class anchor. Upon finding the closest class anchor, in the second stage, distances are computed between the query feature vector and all the embeddings associated with the established class. Distances are then sorted and the closest corresponding items are retrieved. The main advantage of this approach is an improved retrieval time due to the reduced number of required comparisons. A possible disadvantage consists in retrieving wrong items when the retrieved class anchor is not representative for the query image. We present results with our faster alternative in the comparative experiments. 

\noindent
\textbf{Visualizing the embedding space.}
To emphasize the geometric effect of the proposed objective, we have modified the final encoder layer of a ResNet-18 model to output 2D feature vectors without the final non-linearity. The model is trained from scratch on the SVHN dataset by alternatively using the cross-entropy loss, the contrastive loss, and the proposed class anchor margin loss. The resulting embeddings, together with the distribution of the embeddings for each of the models, are presented in Figure~\ref{f:embeds_dist_plot}. For the proposed objective, we have set $m = 4$ and $p = 1$. From the distribution of the embeddings, it can be noted that the proposed objective generates tighter and clearly separable clusters.

\noindent
\textbf{Application to classification tasks.}
An important remark is that we can use the class centers $C$ to apply the proposed system to classification tasks. In this case, the predicted label $\hat{y}$ for a test sample $\mathbf{x}$ can be obtained as follows:
\begin{equation}
    \hat{y} = \arg\min_{j} \| f_{\theta}(\mathbf{x}) - \mathbf{c}_{j} \|.
    \label{e:accuracy}
\end{equation}
We perform additional experiments to demonstrate the application of our class anchor margin loss to classification tasks, with competitive results.



\section{Experiments}

In the experiments, we compare our class anchor margin loss with the cross-entropy and the contrastive learning losses on four datasets, considering both convolutional and transformer models. 

\subsection{Datasets}
We  perform experiments on four datasets: CIFAR-100~\cite{Krizhevsky-TECHREP-2009}, Food-101 \cite{Lukas-ECCV-2014}, SVHN \cite{Netzer-DLUFL-2011}, and Tiny ImageNet~\cite{Russakovsky-IJCV-2015}. 
CIFAR-100 contains 50,000 training images and 10,000 test images belonging to 100 classes.
Food-101 is composed of 101,000 images from 101 food categories. The official split has $750$ training images and $250$ test images per category.
SVHN contains 73,257 digits for training, 26,032 digits for testing.
Tiny ImageNet is a subset of ImageNet-1K, which contains 100,000 training images, 25,000 validation images and 25,000 test images from 200 classes. 

\subsection{Experimental setup}

As underlying neural architectures, we employ three ResNet~\cite{He-CVPR-2016} model variations (ResNet-18, ResNet-50, ResNet-101) and a Swin transformer~\cite{Liu-ICCV-2021} model (Swin-T). We rely on the PyTorch~\cite{r:pytorch} library together with Hydra~\cite{r:hydra} to implement and test the models.

We apply random weight initialization for all models, except for the Swin transformer, which starts from the weights pre-trained on the ImageNet Large Scale Visual Recognition Challenge (ILSVRC) dataset \cite{Russakovsky-IJCV-2015}. 
We employ the Adam~\cite{r:adam} optimizer to train all models, regardless of their architecture. For the residual neural models, we set the learning rate to $10^{-3}$, while for the Swin transformer, we use a learning rate of $10^{-4}$. The residual nets are trained from scratch for 100 epochs, while the Swin transformer is fine-tuned for 30 epochs. For the lighter models (ResNet-18 and ResNet-50), we use a mini-batch size of 512. Due to memory constraints, the mini-batch size is set to 64 for the Swin transformer, and 128 for ResNet-101. Residual models are trained with a linear learning rate decay with a factor of $0.5$. We use a patience of 6 epochs for the full-set experiments, and a patience of 9 epochs for the few-shot experiments. Fine-tuning the Swin transformer does not employ a learning rate decay. Input images are normalized to have all pixel values in the range of $[0, 1]$ by dividing the values by 256. The inputs of the Swin transformer are standardized with the image statistics from ILSVRC~\cite{Russakovsky-IJCV-2015}. 
 
\begin{table*}[!th]
\small
\setlength{\tabcolsep}{.44em}
\begin{center}
{\caption{Retrieval performance levels of ResNet-18, ResNet-50, ResNet-101 and Swin-T models on the CIFAR-100~\cite{Krizhevsky-TECHREP-2009}, Food-101 \cite{Lukas-ECCV-2014}, SVHN \cite{Netzer-DLUFL-2011}, and Tiny ImageNet~\cite{Russakovsky-IJCV-2015} datasets, while comparing the cross-entropy (CE) loss, the contrastive learning (CL) loss, and the class anchor margin (CAM) loss. The best score for each architecture and each metric is highlighted in bold.}\label{table_folds_retrieval}}
\begin{tabular}{ll@{\hspace{10pt}}ccc@{\hspace{10pt}}ccc@{\hspace{10pt}}ccc@{\hspace{10pt}}cccccc}
\toprule
& \multirow{2}{*}{Loss} & \multicolumn{3}{c}{CIFAR-100} & \multicolumn{3}{c}{Food-101} & \multicolumn{3}{c}{SVHN} & \multicolumn{3}{c}{Tiny ImageNet}\\
\rule{0pt}{10pt}
& & mAP & P@20 & P@100 & mAP  & P@20 & P@100 & mAP  & P@20 & P@100 & mAP  & P@20 & P@100\\
\midrule
\multirow{3}{*}{\rotatebox{90}{\scriptsize ResNet-18}}     &  CE    & .249$_{\pm.003}$   & .473$_{\pm.005}$ &  .396$_{\pm.005}$   & .234$_{\pm.002}$ & .547$_{\pm.002}$ & .459$_{\pm.001}$ & .895$_{\pm.013}$    & \textbf{.954$_{\pm.003}$} &  .952$_{\pm.003}$ & .130$_{\pm.001}$   & .340$_{\pm.001}$ & .262$_{\pm.001}$\\
                                                    &  CL     & .220$_{\pm.013}$   & .341$_{\pm.010}$ &  .303$_{\pm.011}$   & .025$_{\pm.001}$ & .042$_{\pm.004}$ & .038$_{\pm.003}$ & .116$_{\pm.000}$   & .115$_{\pm.002}$ & .116$_{\pm.001}$ &  .070$_{\pm.012}$  & .139$_{\pm.017}$ &  .117$_{\pm.016}$\\
                                                    & CAM (ours)    & \textbf{.622$_{\pm.005}$}    & \textbf{.560$_{\pm.007}$} & \textbf{.553$_{\pm.007}$}    & \textbf{.751$_{\pm.001}$} & \textbf{.676$_{\pm.003}$} & \textbf{.669$_{\pm.003}$} & \textbf{.967$_{\pm.001}$}   & \textbf{.954}$_{\pm.000}$ & \textbf{.953$_{\pm.001}$} & \textbf{.508$_{\pm.004}$}   & \textbf{.425$_{\pm.004}$} & \textbf{.418$_{\pm.005}$}\\
\midrule
\multirow{3}{*}{\rotatebox{90}{\scriptsize ResNet-50}}     &  CE     & .211$_{\pm.006}$   & .454$_{\pm.007}$ & .366$_{\pm.006}$   & .158$_{\pm.004}$ & .471$_{\pm.005}$ & .370$_{\pm.005}$ & .909$_{\pm.010}$   & \textbf{.958$_{\pm.001}$} & .956$_{\pm.002}$ & .088$_{\pm.003}$   & .292$_{\pm.006}$ & .209$_{\pm.005}$\\
                                                    &  CL     & .164$_{\pm.016}$  & .271$_{\pm.017}$ & .240$_{\pm.016}$   & .019$_{\pm.000}$ & .030$_{\pm.000}$ & .028$_{\pm.001}$ & .372$_{\pm.310}$   & .447$_{\pm.361}$ & .437$_{\pm.364}$ & .005$_{\pm.000}$    & .008$_{\pm.001}$ & .007$_{\pm.000}$\\
                                                    & CAM (ours)    & \textbf{.640$_{\pm.008}$}   & \textbf{.581$_{\pm.009}$} & \textbf{.578$_{\pm.009}$}   & \textbf{.765$_{\pm.005}$} & \textbf{.697$_{\pm.008}$} & \textbf{.697$_{\pm.009}$} & \textbf{.969$_{\pm.001}$}   & .956$_{\pm.001}$ & \textbf{.957$_{\pm.001}$} & \textbf{.543$_{\pm.003}$}   & \textbf{.472$_{\pm.002}$} & \textbf{.468$_{\pm.002}$}\\
\midrule
\multirow{3}{*}{\rotatebox{90}{\scriptsize ResNet-101}}    &  CE    & .236$_{\pm.009}$   & .482$_{\pm.008}$ & .397$_{\pm.009}$   & .160$_{\pm.003}$ & .479$_{\pm.008}$ & .376$_{\pm.007}$ & .936$_{\pm.003}$   & \textbf{.958$_{\pm.001}$} & \textbf{.957$_{\pm.001}$} & .093$_{\pm.002}$    & .299$_{\pm.002}$ & .216$_{\pm.002}$\\
                                                    &  CL     & .034$_{\pm.028}$   & .069$_{\pm.051}$ & .056$_{\pm.044}$   & .014$_{\pm.002}$ & .018$_{\pm.004}$ & .017$_{\pm.003}$ & .595$_{\pm.242}$   & .700$_{\pm.295}$ & .696$_{\pm.293}$ & .006$_{\pm.001}$    & .007$_{\pm.002}$ & .007$_{\pm.002}$\\
                                                    & CAM (ours)   & \textbf{.629$_{\pm.006}$} & \textbf{.575$_{\pm.008}$} & \textbf{.573$_{\pm.008}$}   & \textbf{.758$_{\pm.007}$} & \textbf{.690$_{\pm.007}$} & \textbf{.693$_{\pm.006}$} & \textbf{.969$_{\pm.001}$}   & .956$_{\pm.001}$ & .956$_{\pm.001}$ & \textbf{.529$_{\pm.005}$}    & \textbf{.458$_{\pm.006}$} & \textbf{.455$_{\pm.006}$}\\
\midrule
\multirow{3}{*}{\rotatebox{90}{\scriptsize Swin-T}}        &  CE     & .661$_{\pm.004}$   & \textbf{.808}$_{\pm.001}$ & .770$_{\pm.001}$   & .617$_{\pm.006}$ & .817$_{\pm.002}$ & .777$_{\pm.003}$ & {.963}$_{\pm.001}$  & \textbf{.968}$_{\pm.000}$ & \textbf{.968}$_{\pm.000}$ & .560$_{\pm.006}$    & .743$_{\pm.001}$ & .707$_{\pm.003}$\\
                                                    &  CL      & .490$_{\pm.010}$   & .629$_{\pm.007}$ & .584$_{\pm.008}$   & .495$_{\pm.004}$ & .667$_{\pm.001}$ & .633$_{\pm.002}$ & .895$_{\pm.000}$   & .928$_{\pm.001}$ & .924$_{\pm.000}$ & .223$_{\pm.002}$   & .397$_{\pm.001}$ & .338$_{\pm.001}$\\
                                                    & CAM (ours)    & \textbf{.808}$_{\pm.004}$   & .794$_{\pm.003}$ & \textbf{.795}$_{\pm.004}$   & \textbf{.873$_{\pm.001}$} & \textbf{.858$_{\pm.001}$} & \textbf{.861$_{\pm.001}$} & \textbf{.967$_{\pm.002}$}   &  .944$_{\pm.001}$ & .946$_{\pm.001}$ & \textbf{.761}$_{\pm.002}$   & \textbf{.745}$_{\pm.003}$ & \textbf{.749}$_{\pm.004}$\\
\bottomrule
\end{tabular}
\end{center}
\end{table*}

\begin{figure}[t]
\centerline{\includegraphics[width=\linewidth]{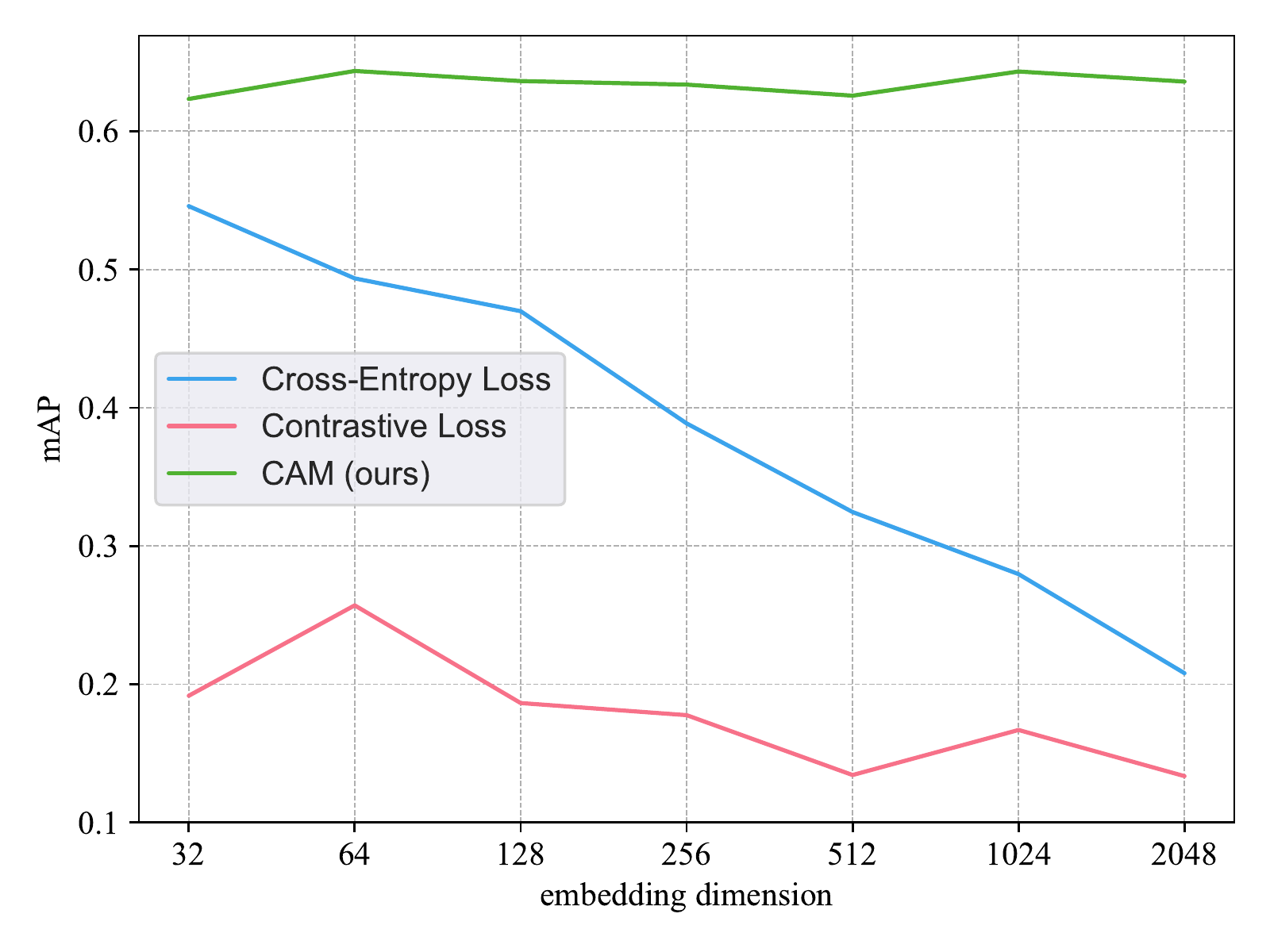}}
\caption{Performance (mAP) of ResNet-50 models with different embedding dimensions from 32 to 2048, trained on CIFAR-100 from scratch. Results are shown for three losses: cross-entropy, contrastive, and class anchor margin (ours). Best viewed in color.}
\label{f:embed_size_perf}
\end{figure}

\begin{figure}[t]
\centerline{\includegraphics[width=\linewidth]{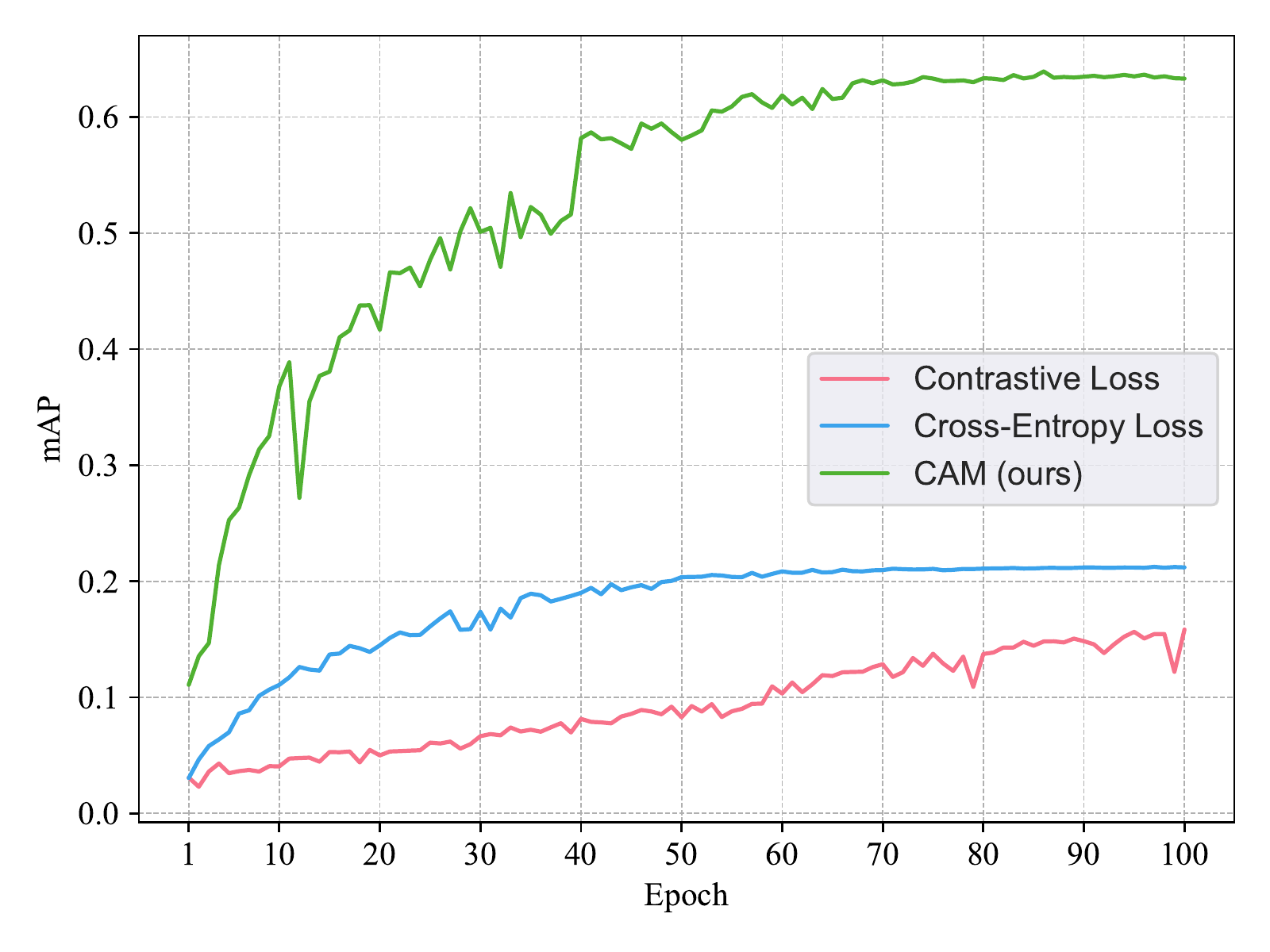}}
\caption{Converge speed of ResNet-50 models trained on CIFAR-100 for 100 epochs using alternative losses: cross-entropy, contrastive, and class anchor margin (ours). Best viewed in color.}
\label{f:map_vs_time}
\end{figure}

We use several data augmentation techniques such as random crop with a padding of 4 pixels, random horizontal flip (except for SVHN, where flipping the digits horizontally does not make sense), color jitter, and random affine transformations (rotations, translations). Moreover, for the Swin transformer, we added the augmentations described in~\cite{r:trivial_aug_wide}. 

For the models optimized either with the cross-entropy loss or our class anchor margin loss, the target metric is the validation accuracy. When using our loss, we set the parameter $m$ for the component $\mathcal{L}_{R}$ to $2$, and the parameter $p$ for the component $\mathcal{L}_{N}$ to $1$, across all datasets and models. Since the models based on the contrastive loss optimize in the feature space, we have used the $1$-nearest neighbors ($1$-NN) accuracy, which is computed for the closest feature vector retrieved from the gallery. 

As evaluation measures for the retrieval experiments, we report the mean Average Precision (mAP) and the precision$@k$ on the test data, where $k \in \{20, 100\}$ is the retrieval rank. For the classification experiments, we use the classification accuracy on the test set. We run each experiment in 5 trials and report the average score and the standard deviation.

\subsection{Retrieval results with full training data}

\noindent
\textbf{Results with various architectures.}
We first evaluate the performance of the ResNet-18, ResNet-50, ResNet-101 and Swin-T models on the CBIR task, while using the entire training set to optimize the respective models with three alternative losses, including our own. The results obtained on the CIFAR-100~\cite{Krizhevsky-TECHREP-2009}, Food-101 \cite{Lukas-ECCV-2014}, SVHN \cite{Netzer-DLUFL-2011}, and Tiny ImageNet~\cite{Russakovsky-IJCV-2015} datasets are reported in Table~\ref{table_folds_retrieval}. First, we observe that our loss function produces better results on the majority of datasets and models. Furthermore, as the rank $k$ increases from 20 to 100, we notice that our loss produces more consistent results, essentially maintaining the performance level as $k$ increases.


\begin{figure*}[t]
\centerline{
\includegraphics[width=\textwidth]{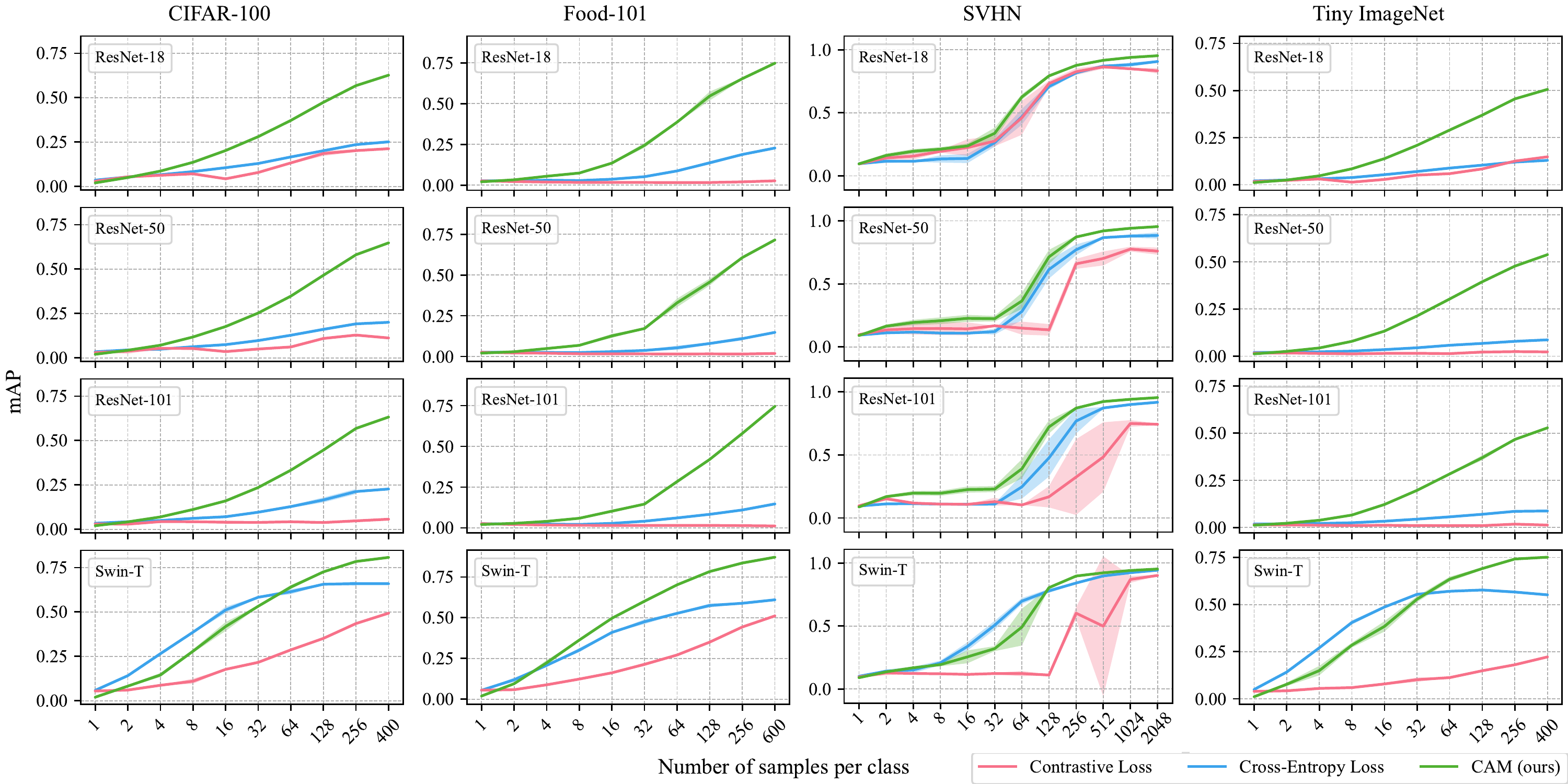}}
\caption{Few-shot retrieval performance (mAP) of four models (ResNet-18, ResNet-50, ResNet-101 and Swin-T) on four datasets (CIFAR-100, Food-101, SVHN and Tiny ImageNet). On each dataset, the results are shown from one sample per class (one-shot learning) to the maximum number of samples per class, by doubling the number of training samples in each trial.}
\label{f:few_shot_performance}
\end{figure*}

\noindent
\textbf{Results with various embedding dimensions.}
We train a ResNet-50 encoder from scratch, where the last encoder layer is modified to output embeddings of various dimensions from the set $\{32, 64, 128, 256, 512, 1024, 2048\}$ on the CIFAR-100 \cite{Krizhevsky-TECHREP-2009} dataset. We illustrate the results with the cross-entropy, contrastive learning and class anchor margin losses in Figure~\ref{f:embed_size_perf}.
We observe that the performance is degrading as the embedding size gets larger  for models trained with cross-entropy or contrastive losses. In contrast, the presented results show that the performance is consistent for the ResNet-50 based on our objective. This indicates that our model is more robust to variations of the embedding space dimension $n$.

\noindent
\textbf{Convergence speed.} We have also monitored the convergence speed of ResNet-50 models, while alternatively optimizing with cross-entropy, contrastive learning and class anchor margin losses. In Figure \ref{f:map_vs_time}, we show how the ResNet-50 models converge over 100 training epochs. The model trained with contrastive loss exhibits a very slow convergence speed. The model trained with cross-entropy achieves its maximum performance faster than the model trained with contrastive-loss, but the former model reaches a plateau after about 60 epochs. The model trained with our CAM loss converges at a faster pace compared with the other models.

\subsection{Few-shot retrieval results}

We next evaluate the neural models on the few-shot object retrieval task. For each dataset, we sample a certain number of training images from each class, starting from one example per class. We gradually increase the number of training samples, doubling the number of training examples per class in each experiment, until we reach the maximum amount of available images. In Figure~\ref{f:few_shot_performance}, we present the corresponding results on CIFAR-100~\cite{Krizhevsky-TECHREP-2009}, Food-101 \cite{Lukas-ECCV-2014}, SVHN \cite{Netzer-DLUFL-2011}, and Tiny ImageNet~\cite{Russakovsky-IJCV-2015}.

For all three ResNet models, our CAM loss leads to better results, once the number of training samples becomes higher or equal to $4$ per class. 
In all cases, the contrastive loss obtains the lowest performance levels, being surpassed by both cross-entropy and class anchor margin losses.
For the Swin-T model, cross-entropy usually leads to better results when the number of samples per class is in the lower range (below $64$). As the number of training samples increases, our loss recovers the gap and even surpasses cross-entropy after a certain point (usually when the number of samples per class is higher than $128$). 
In general, all neural models obtain increasingly better performance levels when more data samples are available for training. With some exceptions, the models trained with our CAM loss achieve the best performance. In summary, we conclude that the class anchor margin loss is suitable for few-shot retrieval, but the number of samples per class should be generally higher than $4$ to obtain optimal performance.

\begin{figure*}[t]
\centerline{
\includegraphics[width=0.96\textwidth]{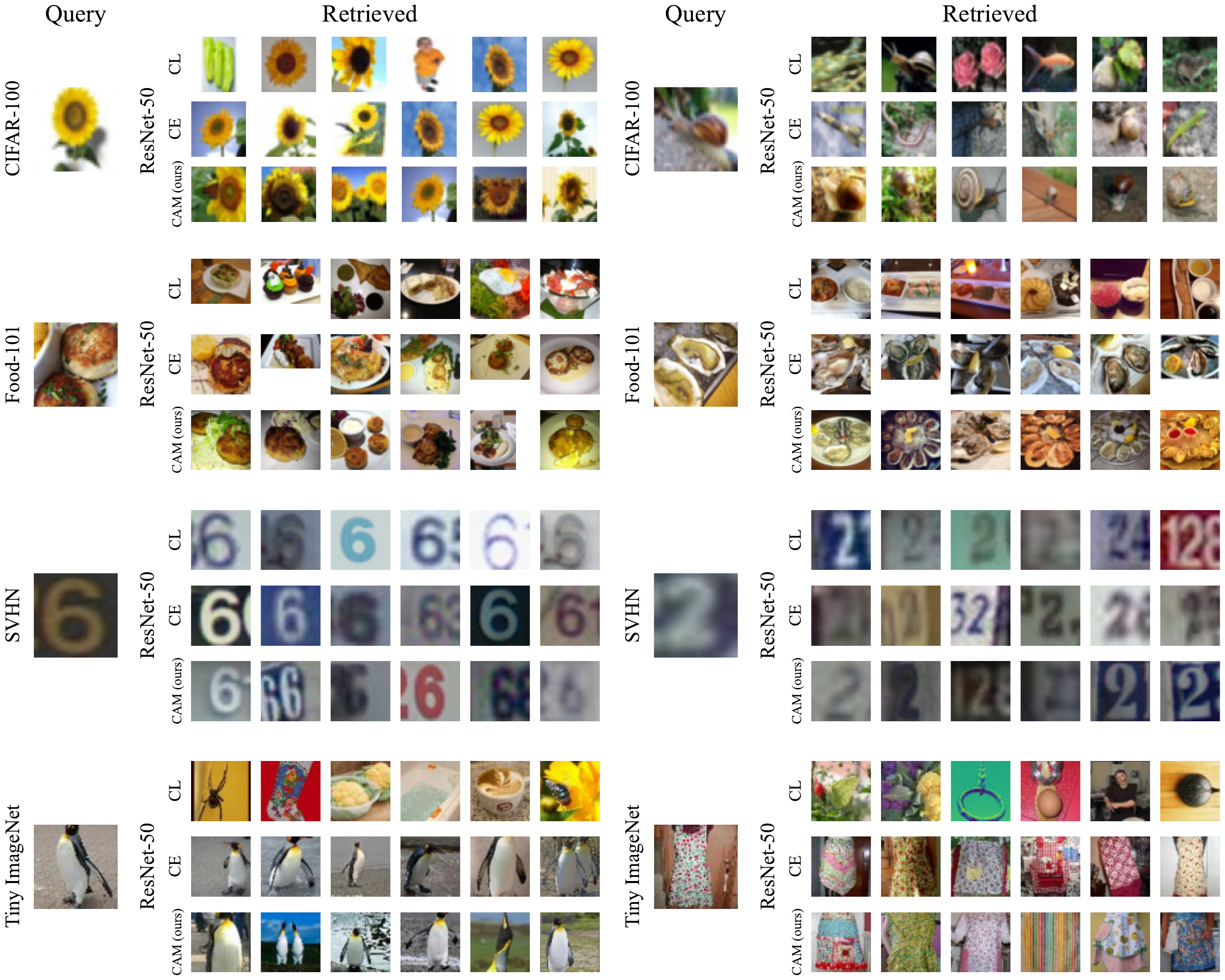}}
\caption{Top 6 retrieved items by a ResNet-50 model trained with one of three losses: cross-entropy (CE), contrastive (CL), and class anchor margin (CAM). We randomly selected two queries per dataset. Best viewed in color.}
\label{f:qualitative_results}
\end{figure*}

\subsection{Qualitative results}

We choose the ResNet-50 model and inspect the retrieved images for each of the three loss functions. In Figure \ref{f:qualitative_results}, we show a set of eight randomly sampled queries from the four datasets (CIFAR-100, Food-101, SVHN, and Tiny ImageNet). The model based on our loss seems to return more consistent results, with the majority of images representing the same object as the query. In contrast, models trained with the other losses can sometimes retrieve items that do not always belong to the query category. Overall, the qualitative results confirm the superiority of our class anchor margin loss.

\begin{table*}[t]
\begin{center}
{\caption{Retrieval performance for brute-force and two-stage retrieval systems on CIFAR-100, Food-101, SVHN and Tiny ImageNet. Results are reported for four models: ResNet-18, ResNet-50, ResNet-101 and Swin-T. The running times are measured on a machine with an Intel Xeon v4 3.0GHz CPU, 256GB of RAM, and an Nvidia GeForce GTX 1080Ti GPU with 11GB of VRAM.}\label{table_two_step_system}}
\begin{tabular}{llcccccccc}
\toprule
\multirow{2}{*}{Model}&\multirow{2}{*}{Method}&\multicolumn{2}{c}{CIFAR-100} & \multicolumn{2}{c}{Food-101} & \multicolumn{2}{c}{SVHN} & \multicolumn{2}{c}{Tiny ImageNet}\\
\cmidrule{3-10}
& & mAP & time (ms) & mAP & time (ms) & mAP & time (ms) & mAP & time (ms)\\
\midrule
\multirow{2}{*}{ResNet-18}      & brute-force   & .476$_{\pm.021}$ & 6.4$_{\pm0.4}$ & .534$_{\pm.003}$   & 9.7$_{\pm0.9}$ & .961$_{\pm.001}$   & 8.8$_{\pm0.9}$ & .328$_{\pm.009}$ & 11.1$_{\pm0.9}$\\
                                & two-stage      & \textbf{.622$_{\pm.005}$}     & \textbf{2.5$_{\pm0.7}$} & \textbf{.751$_{\pm.001}$}      & \textbf{3.2$_{\pm0.8}$} & \textbf{.967$_{\pm.001}$}      & \textbf{3.2$_{\pm0.8}$} & \textbf{.508$_{\pm.004}$} & \textbf{2.5$_{\pm0.7}$}\\
\midrule
\multirow{2}{*}{ResNet-50}      &  brute-force  & .528$_{\pm.016}$  & 14.4$_{\pm1.8}$ & .609$_{\pm.017}$   & 23.2$_{\pm2.0}$ & .963$_{\pm.001}$   & 19.0$_{\pm2.1}$ & .414$_{\pm.003}$ & 25.3$_{\pm3.0}$\\
                                & two-stage      & \textbf{.640$_{\pm.008}$}     & \textbf{5.1$_{\pm0.7}$} & \textbf{.765$_{\pm.005}$}      & \textbf{9.0$_{\pm0.8}$} & \textbf{.969$_{\pm.001}$}      & \textbf{6.8$_{\pm1.2}$} & \textbf{.543$_{\pm.003}$} & \textbf{5.1$_{\pm0.8}$}\\
\midrule
\multirow{2}{*}{ResNet-101}     &  brute-force  & .525$_{\pm.014}$ & 18.8$_{\pm1.8}$ & .618$_{\pm.005}$    & 28.8$_{\pm2.0}$ & .960$_{\pm.003}$   & 23.3$_{\pm2.1}$ & .400$_{\pm.012}$ & 28.6$_{\pm2.5}$\\
                                & two-stage      &  \textbf{.629$_{\pm.006}$}   & \textbf{9.4$_{\pm0.7}$} & \textbf{.758$_{\pm.007}$}       & \textbf{14.6$_{\pm0.8}$} & \textbf{.969$_{\pm.001}$}      & \textbf{11.2$_{\pm1.2}$} & \textbf{.529$_{\pm.005}$} & \textbf{9.5$_{\pm0.7}$}\\
\midrule
\multirow{2}{*}{Swin-T}         &  brute-force  & .762$_{\pm.005}$ & 11.8$_{\pm0.9}$ & .811$_{\pm.002}$    & 14.9$_{\pm0.9}$ & .943$_{\pm.001}$   & 14.3$_{\pm0.9}$ &  .684$_{\pm.008}$ & 17.2$_{\pm1.3}$\\
                                & two-stage      &  \textbf{.808$_{\pm.004}$}   & \textbf{6.9$_{\pm0.8}$} & \textbf{.873$_{\pm.001}$}       & \textbf{7.2$_{\pm0.8}$} & \textbf{.967$_{\pm.002}$}      & \textbf{7.7$_{\pm0.9}$} & \textbf{.761$_{\pm.002}$} & \textbf{6.9$_{\pm0.8}$}\\
\bottomrule
\end{tabular}
\end{center}
\end{table*}

\begin{table}
\begin{center}
{\caption{Classification and retrieval results while ablating the proposed class anchor initialization and various components of our CAM loss. Results are shown for a ResNet-18 model on SVHN.}\label{t:ablation_study}}
\begin{tabular}{ccccc}
\toprule
$\mathcal{L}_{R}$ & $\mathcal{L}_{N}$& Anchors init & Accuracy & mAP \\

\midrule
\multirow{2}{*}{$\times$} & \multirow{2}{*}{$\times$}   &  random           & 19.56\% & 0.116 \\
                                                &       &  base vectors     & 66.74\% & 0.483 \\
\midrule
\multirow{2}{*}{$\times$} & \multirow{2}{*}{\checkmark}   &  random         &  19.58\% & 0.116 \\
                                                &       &  base vectors     &  66.95\% & 0.483 \\
\midrule
\multirow{2}{*}{\checkmark} & \multirow{2}{*}{$\times$}   &  random         & 93.34\% & 0.889 \\
                                                &       &  base vectors     & 94.05\% & 0.901\\
\midrule
\multirow{2}{*}{\checkmark} & \multirow{2}{*}{\checkmark}   &  random       & 93.64\% & 0.892 \\
                                                &       &  base vectors     & 96.41\% & 0.967 \\
\bottomrule
\end{tabular}
\end{center}
\end{table}

\subsection{Ablation study}

\noindent
\textbf{Ablating the loss.}
In Table~\ref{t:ablation_study}, we demonstrate the influence of each additional loss component on the overall performance of ResNet-18 on the SVHN dataset, by ablating the respective components from the proposed objective. We emphasize that the component $\mathcal{L}_{A}$ is mandatory for our objective to work properly. Hence, we only ablate the other loss components, namely $\mathcal{L}_{R}$ and $\mathcal{L}_{N}$.

In addition, we investigate different class center initialization heuristics. As such, we conduct experiments to compare the random initialization of class centers and the base vector initialization. The latter strategy is based on initializing class anchors as scaled base vectors, such that each class center has no intersection with any other class center in the $n$-dimensional sphere of radius $m$, where $n$ is the size of the embedding space. 

As observed in Table~\ref{t:ablation_study}, the class center initialization has a major impact on the overall performance. For each conducted experiment, we notice a significant performance gain for the base vector initialization strategy. Regarding the loss components, we observe that removing both $\mathcal{L}_{R}$ and $\mathcal{L}_{N}$ from the objective leads to very low performance. Adding only the component $\mathcal{L}_{N}$ influences only the overall accuracy, but the mAP is still low, since $\mathcal{L}_{N}$ can only impact each anchor's position with respect to the origin. Adding only the component $\mathcal{L}_{R}$ greatly improves the performance, proving that $\mathcal{L}_{R}$ is crucial for learning the desired task. Using both $\mathcal{L}_{R}$ and $\mathcal{L}_{N}$ further improves the results, justifying the proposed design.



\noindent
\textbf{Ablating the two-stage retrieval system.}
As discussed in Section \ref{s:method}, we employ a two-stage retrieval system to speed up the retrieval process. We hereby ablate the proposed two-stage approach that leverages the class anchors, essentially falling back to the brute-force retrieval process, in which the query is compared with every embedding vector from the database. We compare the ablated (brute-force) retrieval with the two-stage retrieval in Table \ref{table_two_step_system}. Remarkably, we observe that our two-stage retrieval system not only improves the retrieval speed, but also the mAP. In terms of time, the two-stage retrieval improves the speed by a factor ranging between $2\times$ and $3\times$. In terms of performance, the gains are higher than $10\%$ in 9 out of 16 cases. These results illustrate the benefits of our two-stage retrieval process based on leveraging the class anchors.

\begin{table}[t]
\begin{center}
{\caption{Classification accuracy of class anchor margin loss vs.~cross-entropy loss on CIFAR-100, Food-101, SVHN and Tiny ImageNet. Results are reported for two models, ResNet-18 and ResNet-50.}\label{table_folds_acc}}
\begin{tabular}{llcc}
\toprule
\multirow{2}{*}{Dataset}&\multirow{2}{*}{Loss}&\multicolumn{2}{c}{Model}\\
\cmidrule{3-4}
& & ResNet-18 & ResNet-50\\
\midrule
\multirow{2}{*}{CIFAR-100}   &  CE             & 60.53\%$_{\pm0.18}$  & 61.88\%$_{\pm0.66}$ \\       
                            & CAM (ours)       & \textbf{61.94\%$_{\pm0.49}$} & \textbf{63.80\%$_{\pm0.76}$} \\
\midrule
\multirow{2}{*}{Food-101}   &  CE      &   74.66\%$_{\pm0.20}$  & \textbf{76.84\%$_{\pm0.37}$} \\ 
                            & CAM (ours)  & \textbf{75.02\%$_{\pm0.04}$}  & 76.37\%$_{\pm0.52}$ \\
\midrule
\multirow{2}{*}{SVHN}   &  CE   & 96.11\%$_{\pm0.11}$  & 96.30\%$_{\pm0.05}$ \\
                            & CAM (ours) & \textbf{96.41\%$_{\pm0.01}$} & \textbf{96.6\%3$_{\pm0.07}$} \\
\midrule
\multirow{2}{*}{Tiny ImageNet}   &  CE  & \textbf{50.82\%$_{\pm0.33}$} & 52.68\%$_{\pm0.65}$\\
                            & CAM (ours) & 50.67\%$_{\pm0.41}$ & \textbf{54.26\%$_{\pm0.29}$}\\
\bottomrule
\end{tabular}
\end{center}
\end{table}

\subsection{Classification results}

As earlier mentioned, we can leverage the use of the learned class centers and the predictions generated with Eq.~\eqref{e:accuracy} to classify objects into the corresponding categories. We present the classification accuracy rates of the ResNet-18 and ResNet-50 models in Table \ref{table_folds_acc}, while alternating between the cross-entropy and the class anchor margin losses. Our CAM loss provides competitive results, surpassing cross-entropy in 6 out of 8 cases. This confirms that our loss is also suitable for the classification task, even though its performance gains are not as high as for the retrieval task.

\section{Conclusion}

In this paper, we proposed $(i)$ a novel loss function based on class anchors to optimize convolutional networks and transformers for object retrieval in images, as well as $(ii)$ a two-stage retrieval system that leverages the learned class anchors to speed up and increase the performance of the system. We conducted comprehensive experiments using four neural models on four image datasets, demonstrating the benefits of our loss function against conventional losses based on statistical learning and contrastive learning. We also performed ablation studies to showcase the influence of the proposed components, empirically justifying our design choices.

In future work, we aim to extend the applicability of our approach to other data types, beyond images. We also aim to explore new tasks and find out when our loss is likely to outperform the commonly used cross-entropy.

\bibliographystyle{ACM-Reference-Format}
\bibliography{sample-base}


\end{document}